\documentclass{egpubl}
\usepackage{pg2022}

\SpecialIssuePaper         

\usepackage[T1]{fontenc}
\usepackage{dfadobe}  

\usepackage{cite}  
\BibtexOrBiblatex
\electronicVersion
\PrintedOrElectronic
\ifpdf \usepackage[pdftex]{graphicx} \pdfcompresslevel=9
\else \usepackage[dvips]{graphicx} \fi

\usepackage{egweblnk} 
\usepackage{float}
\usepackage[]{graphicx}
\usepackage[abs]{overpic}
\usepackage{amsmath}
\usepackage{amsfonts}
\usepackage{amssymb}
\usepackage{multirow}
\usepackage{subfigure}
\usepackage{epstopdf}
\usepackage{multirow}
\usepackage{color}
\usepackage{bm}
\usepackage{hyperref}
\usepackage[table,xcdraw]{xcolor}

\title[UTOPIC: Uncertainty-aware Overlap Prediction Network for Partial Point Cloud Registration]%
{UTOPIC: Uncertainty-aware Overlap Prediction Network for Partial Point Cloud Registration}

\author[Zhilei Chen et al.]
{\parbox{\textwidth}{\centering Zhilei Chen$^{1}$\orcid{0000-0002-3853-4046}, Honghua Chen$^{1\dag}$, Lina Gong$^{1}$, Xuefeng Yan$^{1}$, Jun Wang$^{1}$, Yanwen Guo$^{3}$, Jing Qin$^{4}$, Mingqiang Wei$^{1,2}$\thanks{Co-corresponding authors: H. Chen and M. Wei.}
        }
        \\
{\parbox{\textwidth}{\centering $^1$School of Computer Science and Technology, Nanjing University of Aeronautics and Astronautics, Nanjing, China\\
$^2$Shenzhen Research Institute, Nanjing University of Aeronautics and Astronautics, Shenzhen, China\\
         $^3$Department of Computer Science, Nanjing University, Nanjing, China\\
         $^4$School of Nursing, Hong Kong Polytechnic University, Hong Kong, China
       }
}
}

\begin{document}

	\teaser{
	 \includegraphics[width=0.95\linewidth, height=0.338\linewidth]{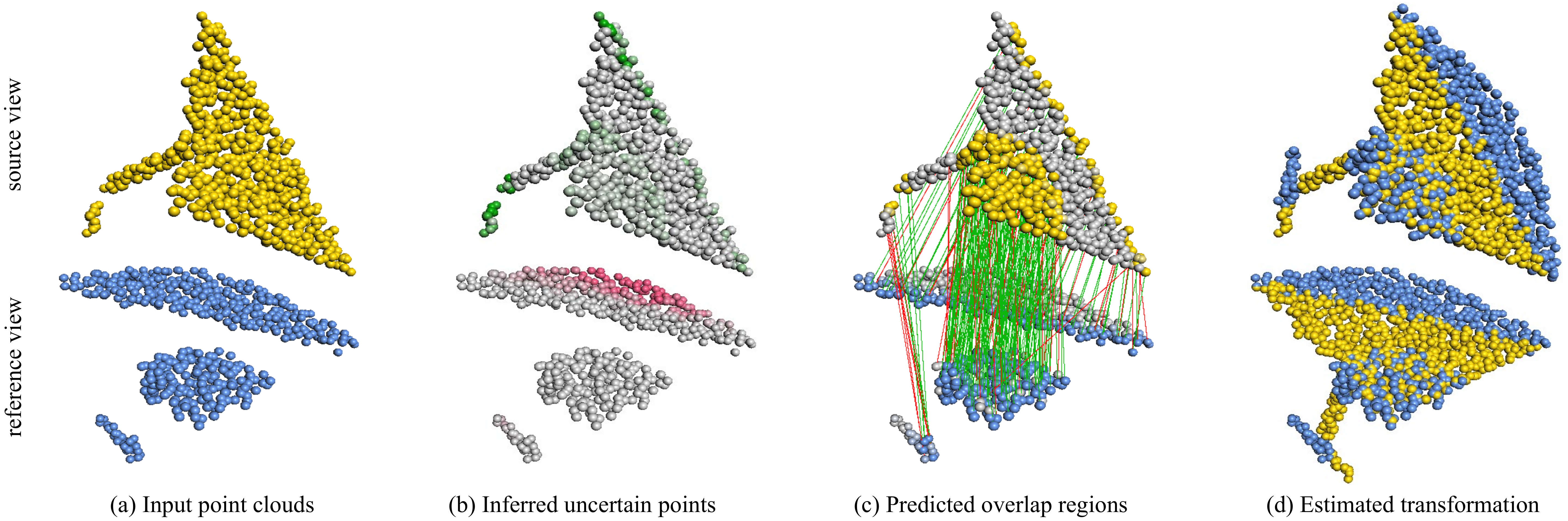}
	 \centering
	  \caption{We present an uncertainty-aware overlap prediction network for partial point cloud registration, which is able to suppress confusing overlapping regions, thus obtaining more reliable overlap scores and more exact dense correspondences for robust registration. The redder or more green points in (b) are the points with higher uncertainties. The colorful points in (c) are points with higher overlap scores. The green lines denote the correct correspondences and the red lines represent the wrong correspondences.} 
	\label{fig:teaser}
	}
	
	\maketitle
	\begin{abstract}
        High-confidence overlap prediction and accurate correspondences are critical for cutting-edge models to align paired point clouds in a partial-to-partial manner. However, there inherently exists uncertainty between the overlapping and non-overlapping regions, which has always been neglected and significantly affects the registration performance. Beyond the current wisdom, we propose a novel uncertainty-aware overlap prediction network, dubbed \emph{UTOPIC}, to tackle the ambiguous overlap prediction problem; to our knowledge, this is the first to explicitly introduce overlap uncertainty to point cloud registration. Moreover, we induce the feature extractor to implicitly perceive the shape knowledge through a completion decoder, and present a geometric relation embedding for Transformer to obtain transformation-invariant geometry-aware feature representations. With the merits of more reliable overlap scores and more precise dense correspondences, \emph{UTOPIC} can achieve stable and accurate registration results, even for the inputs with limited overlapping areas. Extensive quantitative and qualitative experiments on synthetic and real benchmarks demonstrate the superiority of our approach over state-of-the-art methods.

		
	\end{abstract} 

	\section{Introduction}
3D sensors are becoming increasingly available and affordable, including various types of 3D laser scanners, and RGB-D cameras (such as Microsoft Kinect, Intel RealSense, Stereolabs ZED, and Apple Truth Depth Camera) \cite{wu2018automatic, yi2019hierarchical}. This benefits to accurately represent scanned surfaces, detailing the shape, dimensions, and the size of various objects \cite{chen2022repcd}. However, the raw point clouds directly captured by these sensors unavoidably require a registration step to synthesize a complete model or a large-scale scene from multiple partial scans \cite{lu2022transformers}.

The problem of partial point cloud registration has been extensively explored \cite{wang2019prnet,yew2020rpm,ao2021spinnet,wang2021you, huang2021predator, fu2021robust,yew2022regtr, wang2022storm, chen2022imlovenet}, yet not well-solved. Generally, the central of accurate registration lies in two aspects: (i) \emph{localizing reliable overlapping regions} and (ii) \emph{establishing accurate correspondences}. Specifically, one can sample more matching points from the overlapping portions of two scans \cite{huang2021predator, wang2022storm}, or treat the overlap scores as the confidence of correspondences \cite{yew2022regtr}. The larger the overlap ratio of a point cloud pair is, the easier the overlap region is to be exploited to stitch the point cloud pair for a precise alignment. Therefore, faithfully detecting the most reliable overlapping regions is essential for robust partial registration. On the other hand, given accurate overlapping regions, if the learned features are less distinctive, numerous outlier matches may still lead to a poor alignment \cite{ao2021spinnet,wang2021you}. Moreover, there always exists inherent uncertainty between overlapping and non-overlapping regions. For example, points near the boundary between overlapping and non-overlapping regions are quite ambiguous, especially when the overlap ratio is low or there exist noise and/or outliers. Such kind of uncertainty may confuse the prediction of overlapping points and correspondences, thus significantly hindering the performance of registration.

Unlike most of the existing registration methods, we propose to disentangle partial point cloud registration into two sub-goals: (i) to first enhance the feature representation with ample prior shape knowledge and transformation-invariant geometry relation embedding, and (ii) to introduce the uncertainty into point cloud registration to solve the ambiguous overlap prediction problem.

We follow this path and introduce \emph{\textbf{UTOPIC}}, a neural model for partial point cloud registration with \textbf{U}ncer\textbf{T}ainty-aware \textbf{O}verlap \textbf{P}red\textbf{IC}tion. First, based on the observation that if two partial point clouds can be recovered to their complete versions, it will be easier to align them, we employ a completion decoder to implicitly enrich the feature extractor with geometry shape priors. Second, we collaborate the geometric relation embedding of each point cloud with the Transformer model \cite{vaswani2017attention} to obtain more reliable and geometry-aware feature representations. Based on the above learned features, we propose an uncertainty quantification scheme to measure the overlap uncertainty of each point (see from Fig. \ref{fig:teaser} (b)), and then lower the importance of uncertain regions for subsequent overlap prediction and dense matching. Finally, quality rigid transformation parameters are computed, by considering both overlap scores and correspondences (see from Fig. \ref{fig:teaser} (c)) in the SVD solver. Experiments on both synthetic and real-scanned data and detailed analysis show that our approach achieves state-of-the-art performance compared with its competitors.

Our main contributions are as follows:
\begin{itemize}
  \item We design a novel point cloud registration network, which is the first to explore the overlap uncertainty to predict high-confidence overlap and accurate correspondences for robust registration.
  \item We employ a completion decoder to implicitly enrich the feature extractor of prior shape knowledge, which can be used to output more reliable deep features.
  \item We introduce a geometry transformer incorporated with geometric relation embedding, which better captures the global contextual information in both feature and geometry spaces and enhances cross-feature fusion.
  \item We propose an uncertainty quantification scheme, which is capable of alleviating the ambiguity of overlapping regions.
\end{itemize}

	\section{Related Work}
\subsection{Feature-based Methods}
Feature-based point cloud registration methods usually contain two sub-stages: first extracting powerful feature descriptors and then recovering the transformation with robust pose estimators, e.g., RANSAC. Traditional feature-based methods \cite{rusu2009fast, salti2014shot} use handcrafted features to construct descriptors. With the prosperity of deep learning techniques, learning-based descriptors \cite{choy2019fully, huang2021predator, ao2021spinnet, wang2021you} achieve more impressive and substantial improvements. FCGF \cite{choy2019fully} designs a point descriptor based on fully convolutional network and sparse tensor, which is widely used for correspondence search and registration. \cite{ao2021spinnet} achieves rotation invariance by PCA on the neighborhood to find the normal axis. Predator \cite{huang2021predator} introduces an overlap attention module and uses probabilistic sampling mechanism for robust RANSAC, which significantly improves the performance of registration for low-overlap point cloud pairs. In contrast with the descriptors that rely on local reference frame, \cite{wang2021you} achieves the rotation invariance by grouping equivariant feature learning. However, establishing correspondences based on feature descriptors becomes more challenging when the input scans have ambiguous geometry structures. Besides, these methods are time-consuming due to RANSAC-like estimators.

\subsection{Direct Registration Methods}
Direct registration methods estimate the transformation parameters with a neural network in an end-to-end manner. This kind of method can be further divided into two classes, according to whether depending on correspondences.
\begin{figure*}[!htb]
	\centering{{
			\includegraphics[width=1.0\linewidth]{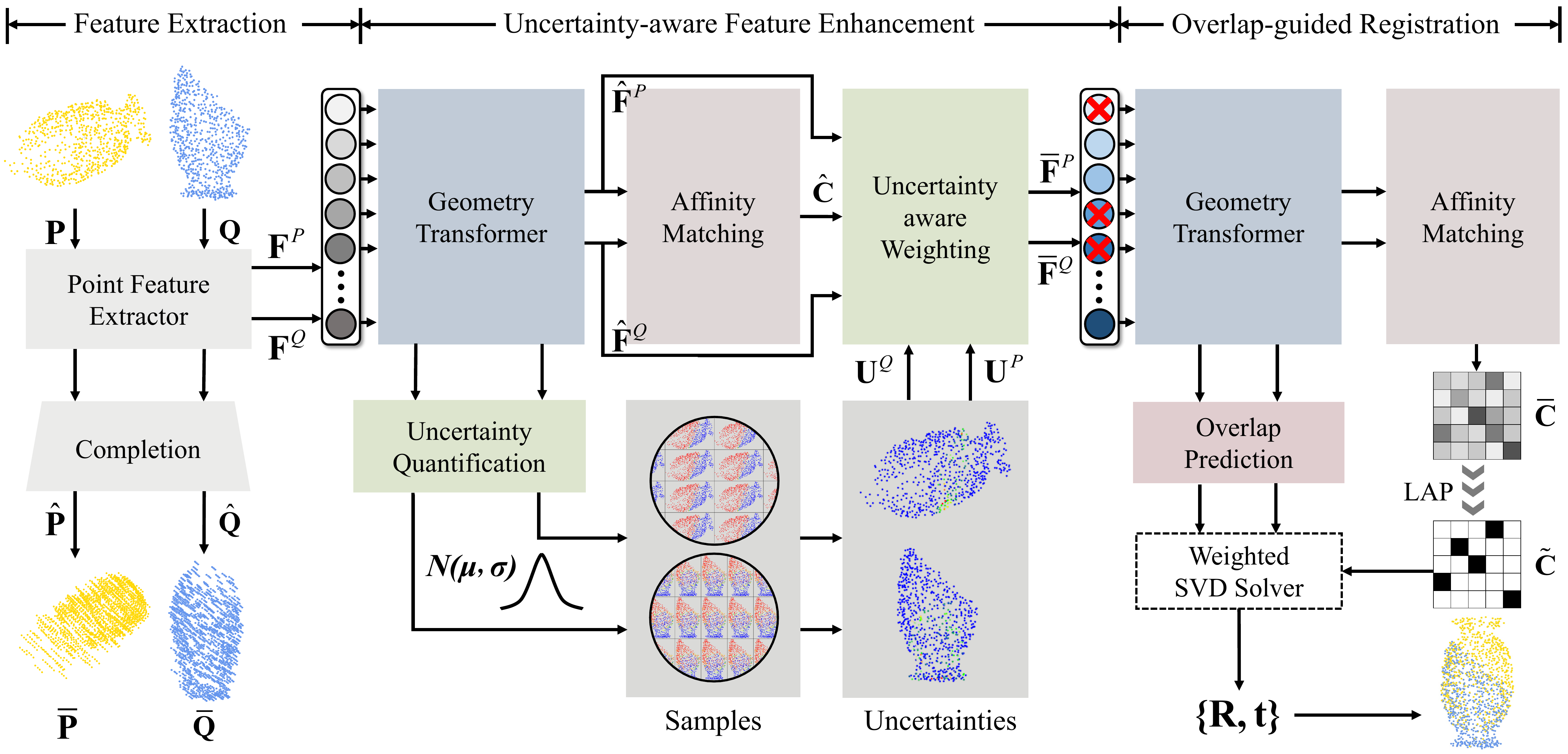}}
		\caption{ \textbf{UTOPIC} contains three parts: Feature Extraction, Uncertainty-aware Feature Enhancement and Overlap-guided Registration.}
		\label{fig:PipeLine}
} \end{figure*}

The correspondence-based methods first extract point-wise features, then calculate the correspondences and finally employ a SVD solver to obtain the transformation. DCP \cite{wang2019deep} mimics the pipeline of ICP \cite{besl1992method}, and employs a transformer to compute soft correspondences and a differentiable SVD layer to extract the transformation. PRNet \cite{wang2019prnet} utilizes a keypoint detector to obtain keypoint-to-keypoint correspondences for partial point cloud registration. To overcome the disadvantage of matching by simply taking the inner product of feature vectors, \cite{li2020iterative} learns to match points based on joint information of the entire geometric features and Euclidean offset for each point pair. \cite{zhang2022end} proposes to learn a partial permutation matching matrix, which does not assign corresponding points to outliers, and implements hard assignment to prevent ambiguity. RPM-Net \cite{yew2020rpm} uses the differentiable Sinkhorn layer and annealing algorithm to get soft assignments of point correspondences from hybrid features. RGM \cite{fu2021robust} introduces a deep graph matching-based framework for registration, which takes both local geometry and graph topological structure into consideration. By contrast, some other methods such as \cite{choy2020deep, pais20203dregnet, bai2021pointdsc, lee2021deep, chen2022sc2} focus more on the removal of outlier correspondences. For example, DGR \cite{choy2020deep} and 3DRegNet \cite{pais20203dregnet} utilize a classifier to predict a confidence value for each correspondence. \cite{bai2021pointdsc} incorporates spatial consistency for pruning outlier correspondences and \cite{chen2022sc2} extends it with a new second-order spatial compatibility. Additionally, \cite{lu2021hregnet} and \cite{qin2022geometric} adopt a coarse-to-fine strategy to better align indoor or large-scale outdoor point clouds. DeepPRO \cite{lee2021deeppro} directly predicts the point-wise location of the aligned point cloud to circumvent overlap prediction or keypoint detection. To deal with the low-overlap point clouds, \cite{wang2022storm} and \cite{yew2022regtr} utilize overlap prediction module to improve the performance of partial registration, but they overlook the inherent uncertainty between overlapping and non-overlapping regions.

The correspondence-free methods extract global features for each point cloud and estimate the transformation with optimization algorithm or regression. PointNetLK \cite{aoki2019pointnetlk} proposes to combine the deep feature networks and the conventional Lucas \& Kanade optimization algorithm to tackle the registration problem. FMR \cite{huang2020feature} presents a feature-metric projection error to align point clouds, which does not need to search the correspondences so that the optimisation speed is fast. \cite{yuan2020deepgmr} leverages a probabilistic registration paradigm by formulating registration as the minimization of KL-divergence between two probability distributions modeled as mixtures of Gaussians. PCRNet \cite{sarode2019pcrnet} simply extracts the global features and regresses the transformation from the combined features. \cite{yan2021consistent} and \cite{li2022unsupervised} solve the  point cloud registration and completion tasks simultaneously, which facilitate each other. In summary, the correspondence-free methods typically use the global features, weakening the importance of local structures. In addition, these methods require time-consuming iterative process to achieve promising results.

\subsection{Uncertainty Modeling}
Generally, there are two typical kinds of uncertainties one can model \cite{der2009aleatory}. Aleatoric uncertainty captures inherent noise from observations, which cannot be reduced even if more data are observed. On the other hand, epistemic uncertainty exists in the model parameters which captures our ignorance about which model generates our collected data. This kind of uncertainty can be explained, if given enough data. Several recent work \cite{zhang2020uc, yang2021uncertainty, liu2022modeling} try to model aleatoric uncertainty by learning the distribution instead of a single fixed uncertainty value. UC-Net \cite{zhang2020uc} proposes probabilistic RGB-D saliency detection network via conditional variational autoencoders to model human annotation uncertainty. \cite{yang2021uncertainty} utilizes a probabilistic representational model in combination with a transformer for camouflaged object detection. \cite{liu2022modeling} proposes a confidence estimation network to model the uncertainty, which is used to dynamically supervise the predicted camouflage maps. Inspired by these works, we propose an uncertainty quantification scheme to capture overlap uncertainty. To the best of our knowledge, this is the first attempt to explicitly introduce overlap uncertainty to point cloud registration. 
	
	\section{Method} 
\subsection{Problem Formulation and Overview}
Given two partial point clouds $\mathbf{P}=\left\{\mathbf{p}_{i} \in \mathbb{R}^{3} \mid i=1, \ldots, N\right\}$ and $\mathbf{Q}=\left\{\mathbf{q}_{j} \in \mathbb{R}^{3} \mid j=1, \ldots, M\right\}$, point cloud registration aims at estimating a rigid transformation $\mathbf{T}=\left\{\mathbf{R, t}\right\}$, where $\mathbf{R} \in \mathbf{SO}(3)$ and $\mathbf{t} \in \mathbb{R}^{3}$. For the simple cases with strict one-to-one correspondences between $\mathbf{P}$ and $\mathbf{Q}$, the transformation parameters can be solved by
\begin{eqnarray} 
	 \min_{\mathbf{R},\mathbf{t}} \sum_{i}^{N} \sum_{j}^{M} \mathbf{C}_{i,j} \left\|\mathbf{R} \mathbf{p}_{i}+\mathbf{t}-\mathbf{q}_{j}\right\|_{2}^{2},
\end{eqnarray}
where $\mathbf{C}$ is the correspondence matrix with $\sum_{j}^{M} \mathbf{C}_{i,j}=1, \forall i$, $\sum_{i}^{N} \mathbf{C}_{i,j}=1, \forall j$, and $\mathbf{C}_{i,j} \in \left\{0,1\right\}^{N \times M}, \forall i,j$. However, for most of the partial overlap scenes, there are no strict one-to-one correspondences. As a result, the correspondence matrix $\mathbf{C}$ cannot satisfy the above constraints. To solve this problem, we also introduce slack variables in $\mathbf{C}$ as \cite{yew2020rpm} to convert inequality constraints back into equality constraints. Specifically, we add an additional row and column to $\mathbf{C}$ for the points that have no correspondences, thus the new matrix $\mathbf{C} \in \mathbb{R}^{(N+1) \times (M+1)}$ can meet above three constraints.

Considering that the high-confidence overlap region is prone to provide more reliable correspondences, we further reformulate the problem of partial point cloud registration as
\begin{eqnarray} \label{eq:formulation}
	\min _{\mathbf{R}, \mathbf{t}} \sum_{i}^{N} \sum_{j}^{M} \widetilde{\mathbf{C}}_{i,j} w_{i,j} \left\|\mathbf{R}\mathbf{p}_{i}+\mathbf{t}-\mathbf{q}_{j}\right\|_{2}^{2},
\end{eqnarray}
where $w_{i,j}$ is the weight computed based on the overlap scores, and $\widetilde{\mathbf{C}}$ is the predicted correspondence matrix. We then propose an end-to-end neural network to predict both the reliable overlap scores and dense correspondences, based on which the rigid transformation is computed via the SVD solver.

Fig.~\ref{fig:PipeLine} shows the overall pipeline of \emph{UTOPIC}, including three key parts: feature extraction, uncertainty-aware feature enhancement, and overlap-guided registration. We first utilize the shared point feature extractor to encode each point in $\mathbf{P}$ and $\mathbf{Q}$, followed by a completion decoder to implicitly enrich the learned feature with the prior shape knowledge. Next, we enhance the feature representation with two sub-modules: a geometry transformer is designed to enhance the global contextual information and promote the cross-feature fusion, and an uncertainty quantification scheme is applied to measure the overlap uncertainty for solving the ambiguous overlap prediction problem. Finally, the geometry transformer and affinity matching are adopted again to detect high-fidelity overlapping regions and accurate dense correspondences, which are then both leveraged to guide the recovery of the alignment transformation. 

\subsection{Feature Extraction}
\begin{figure}[!htb]
	\centering{{
			\includegraphics[width=0.7\linewidth]{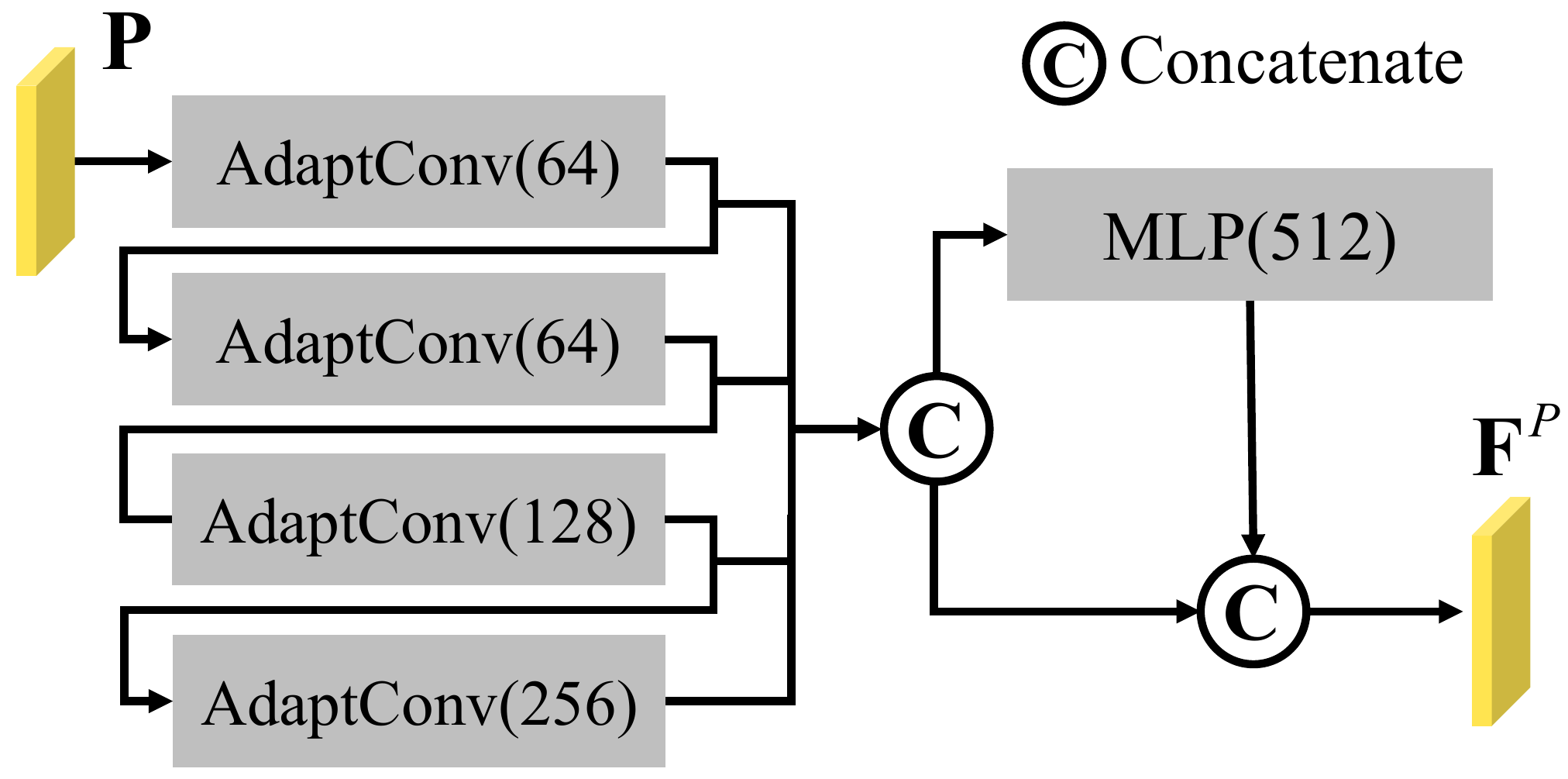}}
		\caption{Detailed architecture of the point feature extractor.}
		\label{fig:PFE}
} \end{figure}

For each input point cloud, we extract multi-level point-wise feature, by using four AdaptConv \cite{zhou2021adaptive} layers (64, 64, 128, 256) followed by an MLP, as shown in Fig.~\ref{fig:PFE}. The learned point-wise feature of $\mathbf{P}$ and $\mathbf{Q}$ are represented as $\mathbf{F}^{P} \in \mathbb{R}^{|\mathbf{P}| \times V}$ and $\mathbf{F}^{Q} \in \mathbb{R}^{|\mathbf{Q}| \times V}$. $V$ is the dimension of the output feature.

It is intuitive that if two partial point clouds can be recovered to their complete versions, it will be easier to align them. Hence, we detach the completion decoder from PCN \cite{yuan2018pcn}, for involving rich geometric shape knowledge into the point-wise feature extractor. The completion decoder produces coarse results $\widehat{\mathbf{P}}$, $\widehat{\mathbf{Q}}$ and fine results $\overline{\mathbf{P}}$, $\overline{\mathbf{Q}}$. During our early experiments, we find that it is not necessary to carefully design a completion model, like \cite{wen2021pmp,xiang2021snowflakenet}.
Even though the completion results are not perfect (see from the left bottom of Fig.~\ref{fig:PipeLine}), it still benefits our registration accuracy (see from the ablation study in Sec.~\ref{sec:abaltion}). We only use the completion decoder to enhance the feature representation. Note also that this completion decoder is only used during the training stage, while removed during testing.

\subsection{Uncertainty-aware Feature Enhancement}
\subsubsection{Geometry Transformer}
\begin{figure}[!htb]
	\centering{{
			\includegraphics[width=1.0\linewidth]{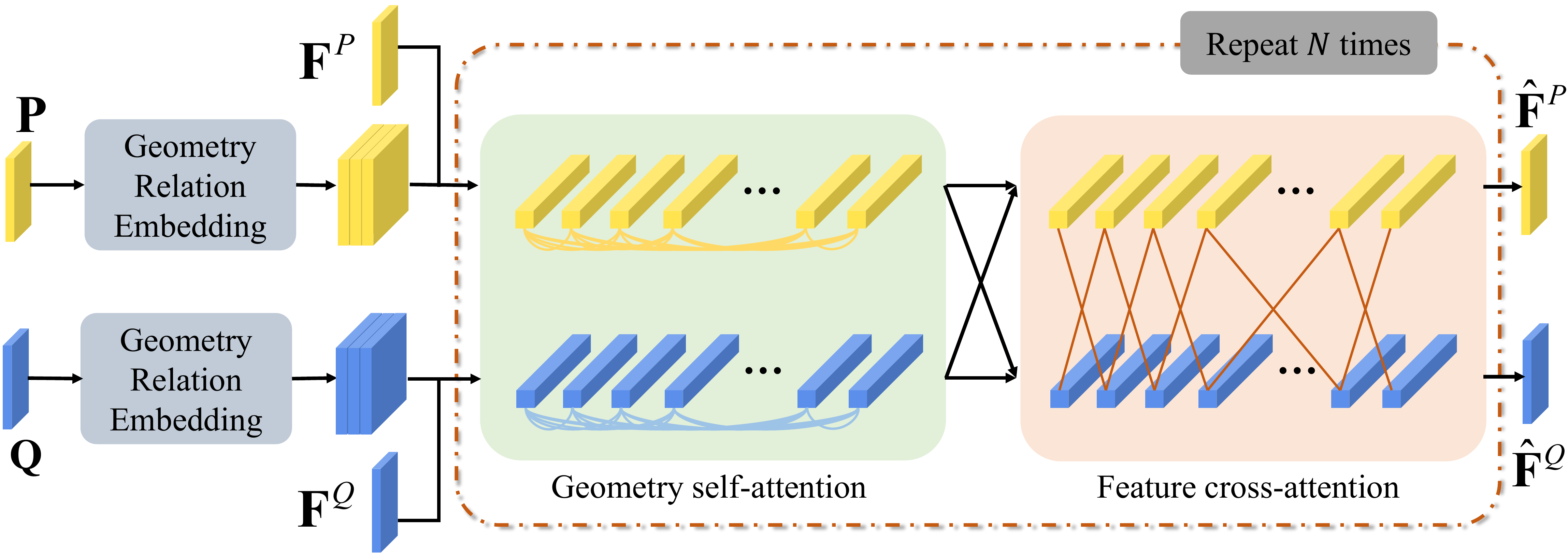}}
		\caption{Detailed architecture of the geometry transformer.}
		\label{fig:GT}
} \end{figure}

Transformer has been proved to be effective for capturing contextual information within a single point cloud or promoting the cross-feature fusion of point cloud pairs \cite{wang2019deep, yu2021cofinet, wang2022storm, yew2022regtr}. Nevertheless, they only feed transformer with deep features and do not take the geometric information into consideration, which makes the learned features less discriminative. \cite{li2022lepard} introduces a position encoding method but merely using the coordinates of points. \cite{qin2022geometric} proposes a geometric relative position embedding, yet with the large memory consumption. To this end, we elaborately design a geometry transformer that not only encodes sufficient geometry information but also occupies relatively low memory (see from Fig.~\ref{fig:GT}). The proposed transformer consists of a geometry self-attention module and a feature cross-attention module. The two modules are alternately iterated for $N$ times to obtain hybrid features $\widehat{\mathbf{F}}^{P} \in \mathbb{R}^{|\mathbf{P}| \times V}$ and $\widehat{\mathbf{F}}^{Q} \in \mathbb{R}^{|\mathbf{Q}| \times V}$.

\begin{figure}[!htb]
	\centering{{
			\includegraphics[width=1.0\linewidth]{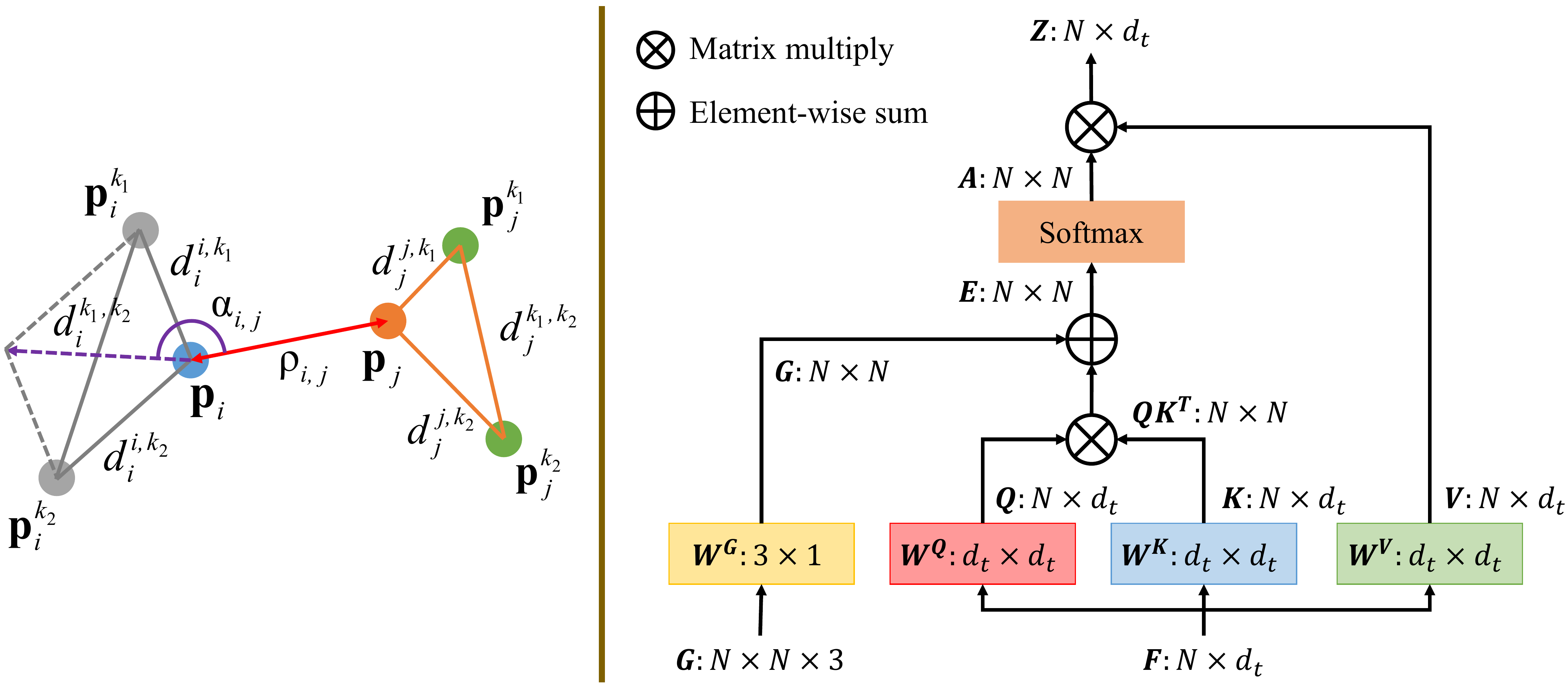}}
		\caption{Left: An illustration of the geometric relation embedding. Right: Computation flow of the geometry self-attention.}
		\label{fig:GRE}
} \end{figure}

\textbf{Geometry self-attention.} The geometry self-attention is designed to learn the global contextual information in both feature and geometry spaces for each point cloud. We describe the computation for $\mathbf{P}$ (the same for $\mathbf{Q}$). Given the features $\mathbf{F} \in \mathbb{R}^{|\mathbf{P}| \times d_{t}}$, where $d_{t}$ is the dimension of the hidden feature, the output features $\mathbf{Z} \in \mathbb{R}^{|\mathbf{P}| \times d_{t}}$ are the weighted sum of all projected input features as
\begin{eqnarray} 
	\mathbf{z}_{i}=\sum_{j=1}^{|\mathbf{P}|} a_{i, j}\left(\mathbf{f}_{j} \mathbf{W}^{V}\right),
\end{eqnarray}
where $a_{i, j}$ is the weight coefficient and computed by a row-wise softmax on the attention scores $e_{i, j}$, which is computed as
\begin{eqnarray} 
	e_{i, j}=\frac{\left(\mathbf{f}_{i} \mathbf{W}^{Q}\right)\left(\mathbf{f}_{j} \mathbf{W}^{K}\right)^{T}+\mathbf{g}_{i, j} \mathbf{W}^{G}}{\sqrt{d_{t}}},
\end{eqnarray}
where $\mathbf{g}_{i, j}$ is the geometric relation embedding,  $\mathbf{W}^{Q}, \mathbf{W}^{K}$, and $\mathbf{W}^{V} \in \mathbb{R}^{d_{t} \times d_{t}}$ are the projection weights for queries, keys and values. $\mathbf{W}^{G} \in \mathbb{R}^{3 \times 1}$ is the projection weights for geometric relation embedding. The right part of Fig. \ref{fig:GRE} shows the computation of the geometry self-attention. Note that \cite{qin2022geometric} also embeds geometry structure knowledge into the transformer, but with the extra space complexity of $O(N^{2} \times d_{t})$. By contrast, our geometry self-attention only costs $O(N^{2} \times 3)$. Usually, the feature dimension $d_{t}$ is far larger than 3, our solution therefore has less space complexity.

\textbf{Geometric relation embedding.} We first define the point triplet as the group of a target point associated with its two nearest neighbors. Inspired by the handcrafted features, we also define a novel geometric relation embedding with point-wise distance, triplet-wise angle and the difference between the perimeter of the local triangles. In detail, given two points $\mathbf{p}_{i}$ and $\mathbf{p}_{j}$, their geometric relation can be described as

(1) \emph{Point-wise Distance.} Point-wise distance is the distance $\rho_{i,j}=\left\|\mathbf{p}_{i}-\mathbf{p}_{j}\right\|_{2}$ between two points in the Euclidean space.

(2) \emph{Triplet-wise Angle.} We search two nearest neighbors $\mathbf{p}_{i}^{k_{1}}$, $\mathbf{p}_{i}^{k_{2}}$ for $\mathbf{p}_{i}$, and form a triplet. Then, we calculate the sum of vector $\mathbf{p}_{i}^{k_{1}}-\mathbf{p}_{i}$ and $\mathbf{p}_{i}^{k_{2}}-\mathbf{p}_{i}$, denoted as $\mathbf{p}_{i}^{k}-\mathbf{p}_{i}$. The triplet-wise angle is computed as $\alpha_{i, j}=\angle\left(\mathbf{p}_{i}^{k}-\mathbf{p}_{i},\mathbf{p}_{j}-\mathbf{p}_{i}\right)$.

(3) \emph{Triangle Perimeter Difference.} We also denote the two nearest neighbors of $\mathbf{p}_{j}$ as $\mathbf{p}_{j}^{k_{1}}$ and $\mathbf{p}_{j}^{k_{2}}$, which form a local triangle. As is shown in the left part of Fig.~\ref{fig:GRE}, we compute the perimeter difference between local triangles as $\eta_{i,j}=\left(d_{i}^{i,k_{1}}+d_{i}^{i,k_{2}}+d_{i}^{k_{1},k_{2}}\right)-\left(d_{j}^{j,k_{1}}+d_{j}^{j,k_{2}}+d_{j}^{k_{1},k_{2}}\right)$, where $d_{i}^{i,k_{1}}=\left\|\mathbf{p}_{i}-\mathbf{p}_{i}^{k_{1}}\right\|_{2}$.

Finally, the geometric relation embedding $\mathbf{g}_{i,j}$ is computed by aggregating the point-wise distance, triplet-wise angle and the triangle perimeter difference as
\begin{eqnarray} 
	\mathbf{g}_{i, j}=CAT[\rho_{i, j}, \alpha_{i, j}, \eta_{i,j}],
\end{eqnarray}
where $CAT[\cdot, \cdot]$ represents the concatenation operation.

\textbf{Feature cross-attention.} Cross-attention is essential for the feature interaction between two point clouds. Given the features $\mathbf{F}^{P}$, $\mathbf{F}^{Q}$ of $\mathbf{P}$, $\mathbf{Q}$, the cross-attention features $\mathbf{Z}^{P}$ of $\mathbf{P}$ is computed as
\begin{eqnarray}
    \mathbf{z}_{i}^{P}=\sum_{j=1}^{|\mathbf{Q}|} a_{i, j}\left(\mathbf{f}_{j}^{Q} \mathbf{W}^{V}\right).
\end{eqnarray}

Similarly, $a_{i, j}$ is computed by a row-wise softmax on the cross-attention scores $e_{i, j}$ as
\begin{eqnarray} 
	e_{i, j}=\frac{\left(\mathbf{f}_{i}^{P} \mathbf{W}^{Q}\right)\left(\mathbf{f}_{j}^{Q} \mathbf{W}^{K}\right)^{T}}{\sqrt{d_{t}}}.
\end{eqnarray}

The cross-attention feature of $\mathbf{Q}$ is computed in the same way. The geometry self-attention module encodes the transformation-invariant geometric relation embedding for each point cloud, while the feature cross-attention module encourages the feature interaction conditioned on each other. Therefore, the final feature representation is more robust and discriminative.

\begin{figure}[!htb]
	\centering{{
			\includegraphics[width=1.0\linewidth]{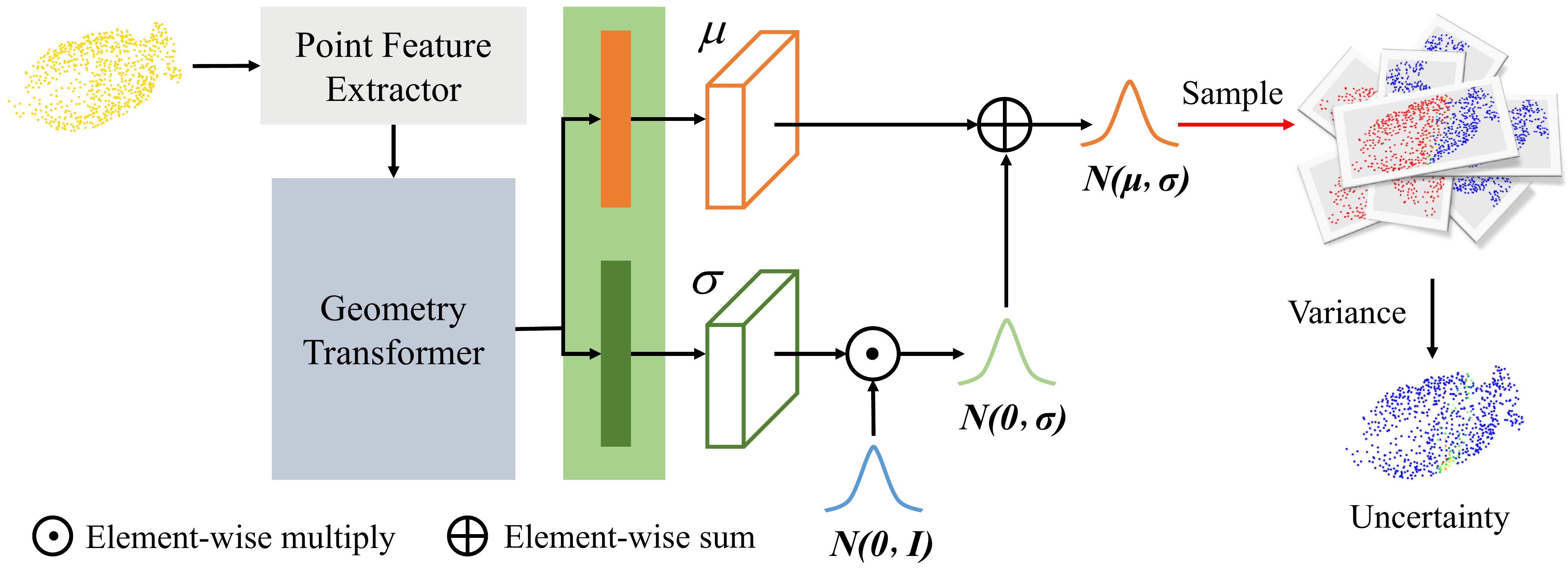}}
		\caption{Overview of the uncertainty quantification module. This module decomposes the sample operation into a trainable part and a random part for end-to-end learning.}
		\label{fig:UQ}
} \end{figure}

\subsubsection{Uncertainty Quantification}
In the partial overlap scenes, it is not trivial to find the one-to-one correspondence. Outlier points may lead to many mismatches. To resolve this problem, recent literature \cite{huang2021predator, xu2021omnet, wang2022storm, yew2022regtr} is devoted to better feature representation and reliable overlap predictions. However, these approaches neglect the uncertainty that inherently exists between overlapping and non-overlapping regions and in some other challenging regions, especially when the overlap ratio is low or there exist noise and outliers. 

Motivated by this issue, we propose a probability-based overlap uncertainty quantification scheme to generate more discriminative features. In detail, for the input point cloud $\mathbf{P}$, we assign each point a Gaussian distribution parameterized by the mean $\mu_{i}$ and the variance $\sigma_{i}$, which are predicted by different MLP layers from $\widehat{\mathbf{F}}^{P}$. The overlap score of $\mathbf{p}_{i}$ can be then sampled from the learned distribution, $o_{i} \sim \mathcal{N}(\mu_{i}, \sigma_{i})$. Following \cite{yang2021uncertainty}, we decompose the sampling operation into two steps (see from Fig.~\ref{fig:UQ}). First, one random sample $\epsilon_{i}$ is obtained from the standard Gaussian distribution $\mathcal{N}(0, I)$, i.e., $\epsilon_{i} \sim \mathcal{N}(0, I)$, and the desired sample is computed by $\mu_{i}+\epsilon_{i} \sigma_{i}$. To measure point-wise overlap uncertainty, we then re-sample $K$ overlap scores for each point from the learned distribution, denoted as $O^{P}=\left\{o^{(1)},...,o^{(K)}\right\}$. According to Bayesian probability theory, we can treat $O^{P}$ as empirical samples from an approximate predictive distribution and measure the uncertainty by calculating the variance as
\begin{eqnarray}
\label{eq:uncertainty}
    \mathbf{U}^{P}=norm(var(O^{P})),
\end{eqnarray}
where $\mathbf{U}^{P} \in \mathbb{R}^{|\mathbf{P}| \times 1}$ represents the overlap uncertainty, $norm(\cdot)$ is the min-max normalization and $var(\cdot)$ is the variance computation. Also, the overlap uncertainty $\mathbf{U}^{Q}$ can be obtained by the same way.

\begin{figure}[!tbh]
	\centering{{
			\includegraphics[width=1.0\linewidth]{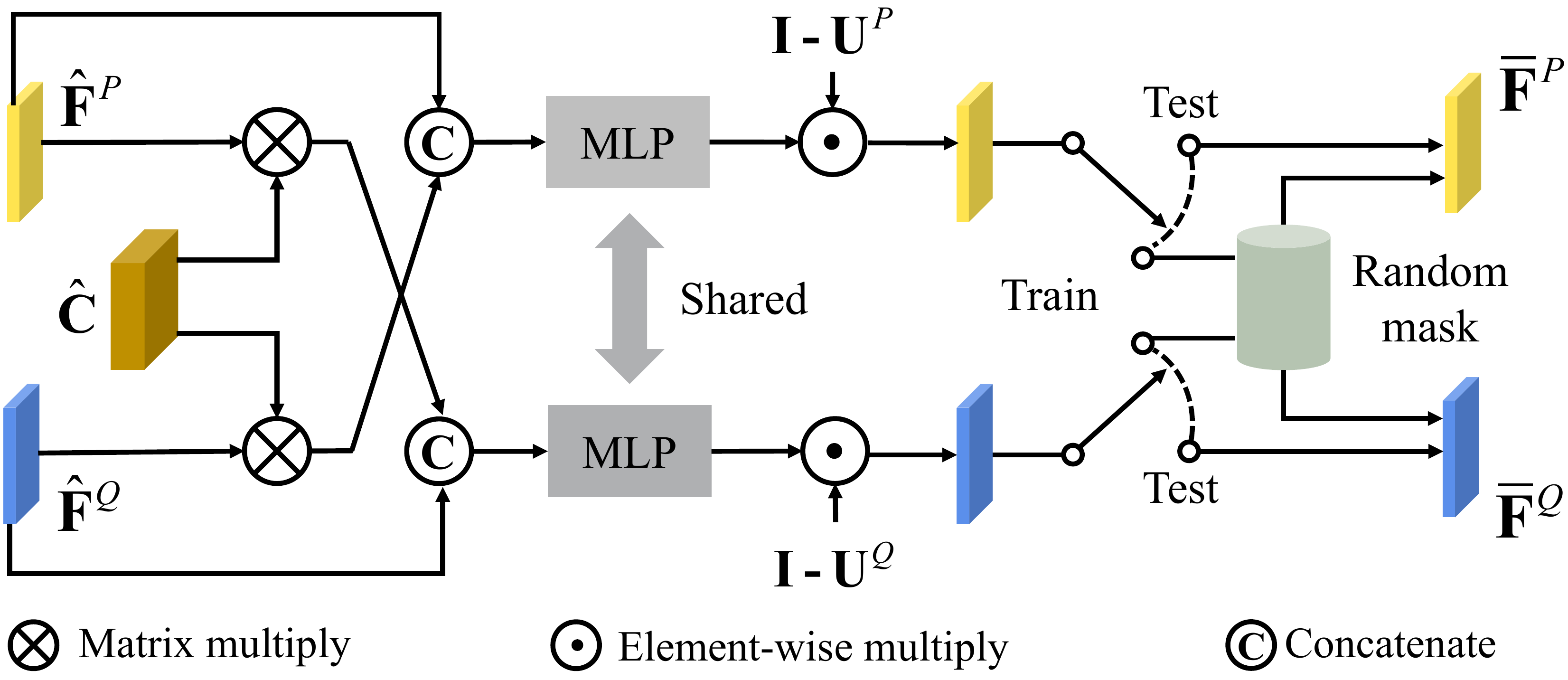}}
		\caption{The details of the uncertainty-aware weighting module.}
		\label{fig:UAW}
} \end{figure}

\subsubsection{Uncertainty-aware Weighting}
Once obtaining the overlap uncertainty, the learned point-wise features are re-weighted to highlight the regions of low uncertainties. As illustrated in Fig.~\ref{fig:UAW}, we take the features $\widehat{\mathbf{F}}^{P}, \widehat{\mathbf{F}}^{Q}$, the correspondence $\widehat{\mathbf{C}}$, and the overlap uncertainty $\mathbf{U}^{P} (\mathbf{U}^{Q})$ as inputs to compute the uncertainty-aware features $\overline{\mathbf{F}}^{P} (\overline{\mathbf{F}}^{Q})$ as
\begin{eqnarray}
    \widetilde{\mathbf{F}}^{P}=MLP(cat[\widehat{\mathbf{F}}^{P}, \widehat{\mathbf{C}} \otimes \widehat{\mathbf{F}}^{Q}]),
\end{eqnarray}
\begin{eqnarray}
    \overline{\mathbf{F}}^{P}=\widetilde{\mathbf{F}}^{P} \odot (I-\mathbf{U}^{P}),
\end{eqnarray}
where $MLP(\cdot)$ denotes a multi-layer perception. 
Although this module is designed to lower the importance of features from the uncertain regions, the network are expected to pay more attention to these regions. We take an uncertainty-based random masking strategy, as demonstrated in the right-most part of Fig.~\ref{fig:UAW}. During training stage, we first assign a random number to each point, and then mask out those features whose associated overlap uncertainty is larger than the random value. The uncertainty-based random masking strategy increases the difficulty and diversity of training samples. As a result, the uncertainty information can be incorporated into the training procedure. This motivates the network to focus more on the learning of these uncertain points. Note that the random masking strategy is removed during testing stage.

\textbf{Affinity matching.} Notably, in the process of the above uncertainty-aware weighting, a soft correspondence $\widehat{\mathbf{C}}$ is required. An affinity matching module is used to generate it. We calculate the similarity matrix of features $\widehat{\mathbf{F}}^{P}$, $\widehat{\mathbf{F}}^{Q}$ in a learnable way instead of simple dot-product of feature vector, which can be represented as
\begin{eqnarray}
    \mathbf{A}_{i, j}=\left(\hat{\mathbf{f}}_{i}^{P}\right)^{T} \mathbf{W}\left(\hat{\mathbf{f}}_{j}^{Q}\right),
\end{eqnarray}
where $\mathbf{W} \in \mathbb{R}^{V \times V}$ is the learnable parameter. For handling outliers and recovering equality constraints of the correspondences matrix, we add a slack variable to row and column of the similarity matrix, and then leverage Sinkhorn \cite{sinkhorn1964relationship} algorithm to calculate the soft correspondence matrix $\widehat{\mathbf{C}}$. More details please refer to \cite{yew2020rpm}.

\subsection{Overlap-guided Registration}
As previously investigated, approaches \cite{huang2021predator, xu2021omnet, wang2022storm, yew2022regtr} present different strategies to improve the performance of registration with overlap scores. For example, Predator \cite{huang2021predator} samples more interest points from overlapping regions for RANSAC. \cite{wang2022storm} applies gumbel softmax to sample the overlap correspondences. OMNet \cite{xu2021omnet} utilizes overlap masks to highlight the points within the overlap regions. We argue that only using the predicted overlap points to estimate the rigid transformation is unreliable for the reason that sparse correspondences from these points may be insufficient and confused due to repetitive structures, noise and outliers \cite{huang2021comprehensive}. Hence, we propose an overlap-guided registration module to tackle this problem.

With the uncertainty-aware features $\overline{\mathbf{F}}^{P}, \overline{\mathbf{F}}^{Q}$, another geometry transformer is adopted to predict the reliable overlap scores, $\widehat{\mathbf{O}}^{P}$ and $\widehat{\mathbf{O}}^{Q}$. Meanwhile, the final soft correspondence $\overline{\mathbf{C}}$ is also obtained by an affinity matching module. We then leverage the linear assignment problem (LAP) solver \cite{jonker1987shortest} to compute the binary hard correspondence $\widetilde{\mathbf{C}}$ from $\overline{\mathbf{C}}$. Based on observation that the correspondence is more likely to be accurate if a pair of corresponding points also both lie in the overlapping region, we add this weight for each correspondence as:
\begin{eqnarray}
    w_{i,j}=\left\{\begin{matrix}
  \frac{\widehat{\mathbf{o}}_{i}^{P} \widehat{\mathbf{o}}_{j}^{Q}}
    {\sum_{i}^{N} \sum_{j}^{M}  \widetilde{\mathbf{C}}_{i,j} \widehat{\mathbf{o}}_{i}^{P} \widehat{\mathbf{o}}_{j}^{Q}}, & \widetilde{\mathbf{C}}_{i,j}=1 \\
    0, &\widetilde{\mathbf{C}}_{i,j}=0
\end{matrix}\right..
\end{eqnarray}

Finally, the rigid transformation is easily computed by SVD, according to Eq.~\ref{eq:formulation}.

\begin{table*}[!htb]
\caption{Quantitative comparison of different methods on ModelNet, ModelLoNet and ScanObjectNN. The best performance is highlighted in \textbf{bold}, and the sub-optimal performance is marked by \underline{underline}.}
\label{tab:Comparison}
\centering{
\begin{tabular}{cccccccc}
\hline
Dataset           & Method            & RMSE(\textbf{R})                                & MAE(\textbf{R})                                 & RMSE(\textbf{t})                                 & MAE(\textbf{t})                                  & Error(\textbf{R})                               & Error(\textbf{t})                                \\ \hline
                           & IDAM \cite{li2020iterative}     & 10.8991                                & 7.4032          & 0.14258          & 0.09009          & 14.2849         & 0.19260          \\
                           & RPM-Net \cite{yew2020rpm}  & 2.5491                                 & 0.8849                                 & 0.01984                                 & 0.00865                                 & 1.6914                                 & 0.01830                                 \\
                           & OMNet \cite{xu2021omnet}    & 8.2406                                 & 5.8538                                 & 0.09341                                 & 0.06161                                 & 11.4084                                & 0.12982                                 \\
                           & RGM \cite{fu2021robust}      & 3.0068                                 & 0.8072                                 & 0.02556                                 & 0.00699                                 & 1.5610                                 & 0.01458                                 \\
                           & Predator \cite{huang2021predator} & {\underline{0.6505}}          & 0.4609                                 & 0.01148                                 & 0.00447                                 & 0.8643                                 & 0.00904                                 \\
                           & REGTR \cite{yew2022regtr}    & 1.1690                                 & {\underline{0.1894}}          & {\underline{0.00434}}          & {\underline{0.00153}}          & {\underline{0.3493}}          & {\underline{0.00306}}          \\
\multirow{-7}{*}{\begin{tabular}[c]{@{}c@{}}ModelNet \\ \cite{wu20153d}\end{tabular}} & Ours     & {\textbf{0.2238}} & {\textbf{0.1322}} & {\textbf{0.00212}} & {\textbf{0.00131}} & {\textbf{0.2464}} & {\textbf{0.00264}} \\ \hline
                           & IDAM \cite{li2020iterative}    & 16.9098                                & 11.7129                                & 0.23770                                 & 0.15980                                 & 22.6078                                & 0.34240                                 \\
                           & RPM-Net \cite{yew2020rpm}  & 6.6189                                 & 2.0453                                 & 0.04813                                 & 0.01827                                 & 3.8514                                 & 0.03835                                 \\
                           & OMNet \cite{xu2021omnet}   & 10.2776                                & 7.7237                                 & 0.12428                                 & 0.08294                                 & 15.0794                                & 0.17329                                 \\
                           & RGM \cite{fu2021robust}     & 17.8005                                & 5.5261                                 & 0.10970                                 & 0.04025                                 & 9.8361                                 & 0.08408                                 \\
                           & Predator \cite{huang2021predator} & 16.0141         & 3.3206                                 & 0.12510                                 & 0.03281                                 & 5.8900                                 & 0.07673                                 \\
                           & REGTR \cite{yew2022regtr}   & {\underline{3.4594}}         & {\underline{1.3427}}         & {\underline{0.03216}}          & {\underline{0.01254}}          & {\underline{2.5753}}          & {\underline{0.02617}}          \\
\multirow{-7}{*}{\begin{tabular}[c]{@{}c@{}}ModelLoNet \\ \cite{wu20153d, huang2021predator}\end{tabular}}     & Ours     & {\textbf{2.5463}} & {\textbf{0.8588}} & {\textbf{0.02098}} & {\textbf{0.00765}} & {\textbf{1.5934}} & {\textbf{0.01555}} \\ \hline
                           & IDAM \cite{li2020iterative}    & 15.9602                                & 8.0429                                 & 0.13187                                 & 0.07255                                 & 15.2581                                & 0.15084                                 \\
                           & RPM-Net \cite{yew2020rpm} & 8.1251                                 & 2.8829                                 & 0.08984                                 & 0.03314                                 & 5.4576                                 & 0.07384                                 \\
                           & OMNet \cite{xu2021omnet}   & 10.5122                                & 7.9538                                 & 0.11367                                 & 0.07984                                 & 15.2959                                & 0.16384                                 \\
                           & RGM \cite{fu2021robust}     & 12.6069                                & 3.7596                                 & 0.08729                                 & 0.02740                                 & 7.0376                                 & 0.05634                                 \\
                           & Predator \cite{huang2021predator} & 12.6050                                & 4.8842                                 & 0.12480                                 & 0.04899                                 & 9.4936                                 & 0.10070                                 \\
                           & REGTR \cite{yew2022regtr}   & {\textbf{2.1477}} & {\underline{0.7448}}          & {\underline{0.02268}}          & {\underline{0.00614}}          & {\underline{1.3842}}          & {\underline{0.01247}}          \\
\multirow{-7}{*}{\begin{tabular}[c]{@{}c@{}}ScanObjectNN \\ \cite{uy2019revisiting}\end{tabular}}     & Ours     & {\underline{3.6790}}          & {\textbf{0.2633}} & {\textbf{0.01394}} & {\textbf{0.00179}} & {\textbf{0.5601}} & {\textbf{0.00366}} \\ \hline
\end{tabular}}
\end{table*}

\subsection{Loss Function}
The proposed \emph{UTOPIC} is trained in an end-to-end way without iterative steps, using four loss terms $\mathcal{L}=\mathcal{L}_{r}+\mathcal{L}_{o}+\mathcal{L}_{u}+\mathcal{L}_{c}$. 

\textbf{Registration loss.} We adopt the cross entropy loss between the predicted soft correspondence $\overline{\mathbf{C}}$ and the ground truth correspondence $\mathbf{C}$ to train our model. The formula is as
\begin{eqnarray} 
	\mathcal{L}_{r}=-\sum_{i}^{N} \sum_{j}^{M}\left(\mathbf{C}_{i, j} \log \overline{\mathbf{C}}_{i, j}+\left(1-\mathbf{C}_{i, j}\right) \log \left(1-\overline{\mathbf{C}}_{i, j}\right)\right).
\end{eqnarray}

The ground-truth correspondence $\mathbf{C}$ is computed by mutual nearest neighbour searching as
\begin{eqnarray}
    \mathbf{C}_{i,j}=\left\{\begin{matrix}
  1, & \begin{matrix}
\mathcal{NN}\left(\mathbf{q}_{j} , \mathcal{T}(\mathbf{P})\right)=\mathbf{p}_{i}, \mathcal{NN}\left(\mathcal{T}(\mathbf{p}_{i}), \mathbf{Q}\right)=\mathbf{q}_{j} \\
\text{and } \left\|\mathcal{T}(\mathbf{p}_{i})-\mathbf{q}_{j}\right\|_{2}<r
\end{matrix} \\
  0, & \text {otherwise}
\end{matrix}\right.,
\end{eqnarray}
where $\mathcal{NN}(\cdot)$ denotes the spatial nearest neighbor, $\mathcal{T}(\cdot)$ is the operation of the ground-truth rigid transformation from $\mathbf{P}$ to $\mathbf{Q}$, and $r$ denotes the distance threshold.

\textbf{Overlap loss.} The task of predicting overlap scores is regarded as a binary classification problem and supervised as
\begin{eqnarray}
    \mathcal{L}_{o}^{P}=\frac{1}{|\mathbf{P}|} \sum_{i=1}^{|\mathbf{P}|} \left(\bar{\mathbf{o}}_{i}^{P} \log \left(\widehat{\mathbf{o}}_{i}^{P}\right)+\left(1-\bar{\mathbf{o}}_{i}^{P}\right) \log \left(1-\widehat{\mathbf{o}}_{i}^{P}\right)\right),
\end{eqnarray}
where $\bar{\mathbf{O}}^{P}$ denotes the ground-truth overlap of $\mathbf{P}$ as
\begin{eqnarray}
    \bar{\mathbf{o}}_{i}^{P}=\left\{\begin{matrix}
    1, & \sum_{j}^{M} \mathbf{C}_{i,j}=1 \\
    0, & \text{otherwise}
\end{matrix}\right..
\end{eqnarray}

The overlap loss $\mathcal{L}_{o}^{Q}$ can be computed in the same way. We then get a total overlap loss: $\mathcal{L}_{o}=\mathcal{L}_{o}^{P}+\mathcal{L}_{o}^{Q}$.

\textbf{Uncertainty loss.} To train the uncertainty quantification module, we define the uncertainty loss $\mathcal{L}_{u}$, which is a weighted combination of a standard binary cross entropy (BCE) loss and a Kullback-Leibler (KL) divergence as
\begin{eqnarray}
    \mathcal{L}_{u}^{P}=\lambda \cdot \mathcal{L}_{BCE}\left(o^{(k)}, \bar{\mathbf{O}}^{P}\right)+\eta \cdot KL(\mathcal{N}(\mu, \sigma) \| \mathcal{N}(0, I)),
\end{eqnarray}
where $o^{(k)}$ denotes one example drawn from the overlap distribution, which is used to boost the diversity. $\lambda$ and $\eta$ are balance factors. The final uncertainty loss is $\mathcal{L}_{u}=\mathcal{L}_{u}^{P}+\mathcal{L}_{u}^{Q}$.

\textbf{Completion loss.} Following PCN \cite{yuan2018pcn}, we use the chamfer distance (CD) to supervise the coarse and fine completion results as
\begin{eqnarray}
    \mathcal{L}_{c}^{P}=CD(\widehat{\mathbf{P}}, \widetilde{\mathbf{P}})+CD(\overline{\mathbf{P}}, \widetilde{\mathbf{P}}),
\end{eqnarray}
where $\widetilde{\mathbf{P}}$ is the ground truth completion result. The total completion loss can be computed as $\mathcal{L}_{c}=\mathcal{L}_{c}^{P}+\mathcal{L}_{c}^{Q}$
	
	\begin{figure*}[!htb]
	\centering{{
			\includegraphics[width=1.0\linewidth]{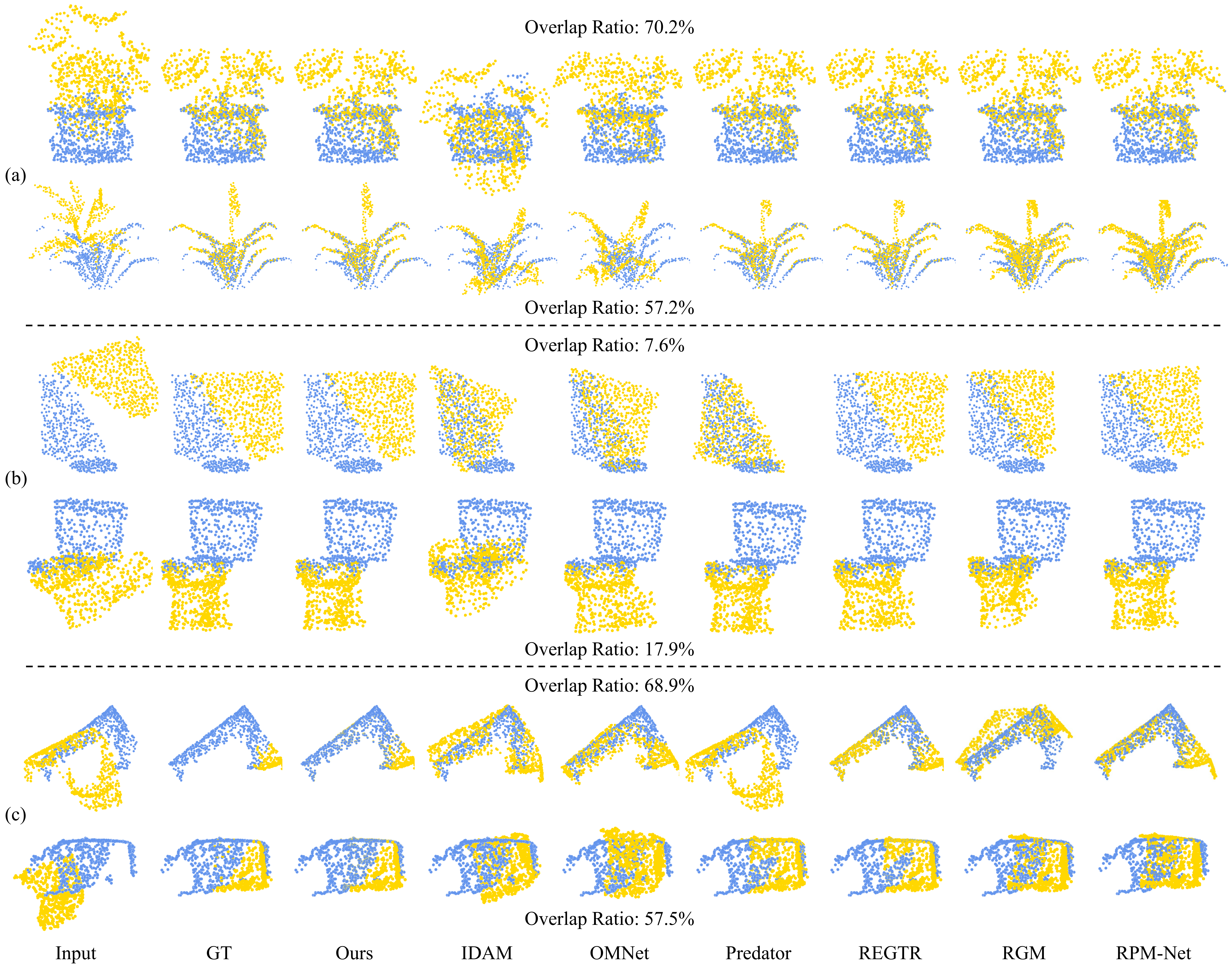}}
		\caption{Comparison of registration results of six SOTA methods and \textbf{UTOPIC} on (a) ModelNet, (b) ModelLoNet, and (c) ScanObjectNN.}
		\label{fig:Comparison}
} \end{figure*}

\section{Experiment}
\textbf{Datasets.} We evaluate our approach on three datasets: ModelNet, ModelLoNet, and ScanObjectNN. 

(i) ModelNet is generated from ModelNet40 \cite{wu20153d}, which contains synthetic point clouds sampled from 12,311 CAD models of 40 different categories. We sample 1024 points for each point cloud, and then adopt the same cropping setting in \cite{yew2020rpm} to simulate the partial-overlap point cloud pairs. Then, one of the point cloud is transformed by a random rotation in the range of $[0, 45]^{\circ}$ and a random translation $\mathbf{t}$ in the range of $[-0.5, 0.5]$ along each axis. Gaussian noise sampled from $\mathcal{N}(0, 0.01)$ and clipped to $[-0.05, 0.05]$ is also added to each coordinate of the points in the clean point clouds. Finally, each point cloud is shuffled to reorder all points. To evaluate the ability of our network to generalize to different shapes, we use the first 20 categories for training (5,112 samples) and validation (1,202 samples) respectively, and the remaining for testing (1,266 samples). 

(ii) ModelLoNet is also generated from ModelNet40. The only difference from ModelNet is that ModelLoNet has an average overlap ratio of 53.6\%, while ModelNet is 73.5\%.

(iii) ScanObjectNN \cite{uy2019revisiting} is a real-world point cloud dataset based on scanned indoor scene data. We use the testing part of ScanObjectNN containing 581 samples. We do not retrain all compared network models. This dataset is only used for testing.

\textbf{Evaluation metrics.} For the consistency with prior work \cite{wang2019deep}, we use the anisotropic metrics of Root Mean Square Error (RMSE) and Mean Average Error (MAE) over Euler angles and translations. Besides, we evaluate the isotropic error for rotation and translation proposed in \cite{yew2020rpm}. All angles are in degrees.

\textbf{Comparison methods.} We compare our method with six latest learning-based registration approaches \cite{li2020iterative, yew2020rpm, xu2021omnet, fu2021robust, huang2021predator, yew2022regtr}. OMNet \cite{xu2021omnet}, Predator \cite{huang2021predator} and REGTR \cite{yew2022regtr} also utilize the overlap prediction. For all compared methods, we use their released public code and follow the same setting in their original papers to retrain the networks by our prepared training data.

\textbf{Implementation details.}
All experiments are conducted on a single Nvidia RTX 2080Ti. We train \emph{UTOPIC} for 200 epochs with a batch size of 4. The SGD optimizer is used with the initial learning rate of 0.001. The sampling number $K$ is set to 50. The balance factors $\lambda, \eta$ are 0.5 and 0.1. The repeat number $N$ in geometry transformer is set to 3, and the distance threshold $r$ is 0.075. The code is available at \href{https://github.com/ZhileiChen99/UTOPIC}{https://github.com/ZhileiChen99/UTOPIC}.

\subsection{Evaluation on Synthetic ModelNet and ModelLoNet}
We quantitatively evaluate the effectiveness of our method on two synthetic
datasets. As reported in Tab.~\ref{tab:Comparison}, our approach achieves the smallest registration errors over all metrics compared with other approaches. REGTR \cite{yew2022regtr} and Predator \cite{huang2021predator} also have impressive results thanks to their overlap prediction mechanisms. When evaluating on the dataset with lower overlap ratio, it is obvious that the performance of most methods degrades dramatically except our approach. In terms of the mean absolute error of rotation in Tab.~\ref{tab:Comparison}, the margin between \emph{UTOPIC} and REGTR \cite{yew2022regtr} on the ModelLoNet is even ten times larger than that on the ModelNet. Fig.~\ref{fig:Comparison} visualizes several cases that are unseen shapes for all neural networks. We can observe that \emph{UTOPIC} achieves the most satisfactory registration results for partial point clouds of noise and sampling irregularity. We can also observe from Fig.~\ref{fig:Comparison} (b) that, our method still produces quality alignments, even though the two input pairs contain rather limited overlapping parts.

\begin{figure}[!htb]
	\centering{{
			\includegraphics[width=0.95\linewidth]{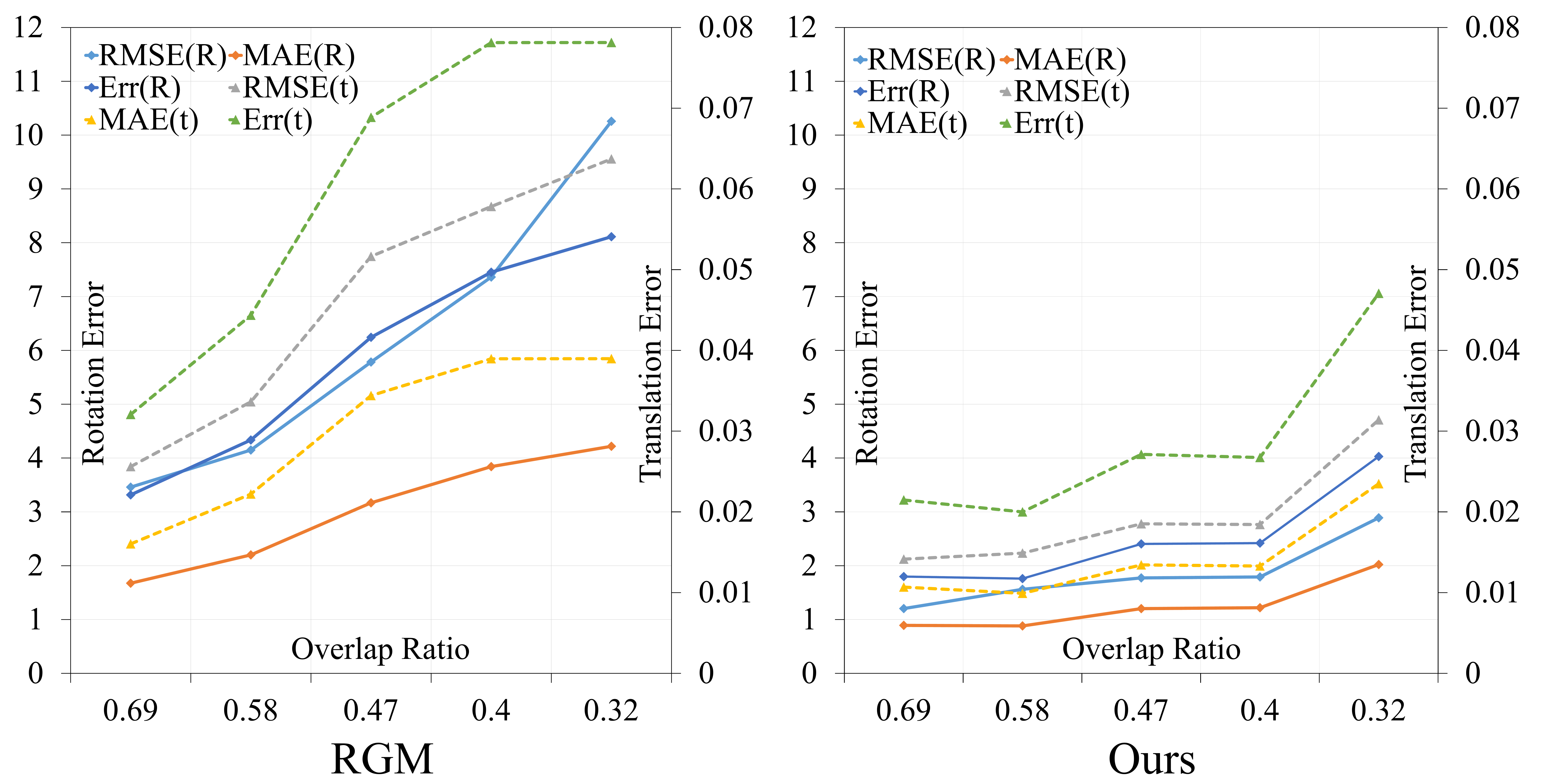}}
		\caption{Error statistics of RGM \cite{fu2021robust} and our method under different overlap ratios.}
		\label{fig:PC}
} \end{figure}

\subsection{Evaluation on ScanObjectNN}
We also evaluate the generalization ability of our method on the ScanObjectNN dataset. The employed models for comparison are all trained on the ModelNet. As shown in the Tab.~\ref{tab:Comparison}, the proposed \emph{UTOPIC} produces almost the best results, which demonstrates a better generalization ability to real-scanned data. Nevertheless, the state-of-the-art methods like \cite{huang2021predator, fu2021robust} are less robust to the data domain gap. Qualitative comparisons are demonstrated in Fig.~\ref{fig:Comparison}, which are consistent with the quantitative statistics.

\begin{table}[!htb]
\caption{Ablation studies on ModelNet. \textbf{O}, \textbf{C}, \textbf{U} and \textbf{G} represent models with overlap, completion, uncertainty and geometric relation embedding. The best performance is highlighted in \textbf{bold}.}
\label{tab:Ablation}
\centering{
\scalebox{0.68}{
\begin{tabular}{cccccccccc}
\hline
O & C & U & G & RMSE(\textbf{R})                       & MAE(\textbf{R})                        & RMSE(\textbf{t})                        & MAE(\textbf{t})                         & Error(\textbf{R})                      & Error(\textbf{t})                       \\ \hline
  &   &   &   &   {0.3682} & {0.1907} & {0.00347} & {0.00195} & {0.3495} & {0.00397} \\
$\surd$ &   &   &   &   {0.2837} & {0.1481} & {0.00241} & {0.00149} & {0.2761} & {0.00300} \\
$\surd$ & $\surd$ &   &   &   {0.2735} & {0.1470} & {0.00238} & {0.00138} & {0.2689} & {0.00277} \\
$\surd$ & $\surd$ & $\surd$ &   & {0.2691} & {0.1404} & {0.00237} & {0.00136} & {0.2600} & {0.00273} \\
$\surd$ & $\surd$ & $\surd$ & $\surd$ & \textbf{0.2238} & \textbf{0.1322} & \textbf{ 0.00212} & \textbf{0.00131} & \textbf{0.2464} & \textbf{0.00264} \\ \hline
\end{tabular}}}
\end{table}

\subsection{Analysis}\label{sec:abaltion}
In this section, we conduct various specified experiments to analyze the main ideas of \emph{UTOPIC}.

\textbf{Ablation study.} We perform ablation studies on the ModelNet dataset for a better understanding of our four main components: (i) the overlap-guided weights (\textbf{O}), (ii) the completion decoder (\textbf{C}), (iii) the uncertainty quantification (\textbf{U}), and (iv) the geometric relation embedding (\textbf{G}). The detailed results are recorded in Tab.~\ref{tab:Ablation}. Clearly, our full pipeline performs the best on all metrics. Removing any component degrades the overall performance, suggesting that all components are beneficial for accurate registration results. In particular, we observe great improvements for the variant with the overlap-guided weights. This implies that overlap prediction is beneficial for partial point cloud registration. This observation also implies that the importance of uncertainty quantification for addressing the ambiguous overlap prediction problem.

\textbf{Different overlap ratios.} Although the comparisons on ModelLoNet have indicated the superiority of our \emph{UTOPIC} for low-overlap point cloud pairs, it is still meaningful to analyse the variation tendency of the performance when overlap ratio decreases gradually. We compare the performance of our approach with RGM \cite{fu2021robust} on ModelNet40. We use the same crop setting of PRNet \cite{wang2019prnet}. The number of points is set to 768, 700, 640, 600 and 560 to generate point clouds with approximate overlap ratios of 0.69, 0.58, 0.47, 0.40, and 0.32, respectively. Fig.~\ref{fig:PC} shows the registration errors for different overlap ratios. As observed, our approach is very stable until the overlap ratio decreases to 0.32, but still better than RGM \cite{fu2021robust}.

\begin{figure}[!htb]
	\centering{{
			\includegraphics[width=0.95\linewidth]{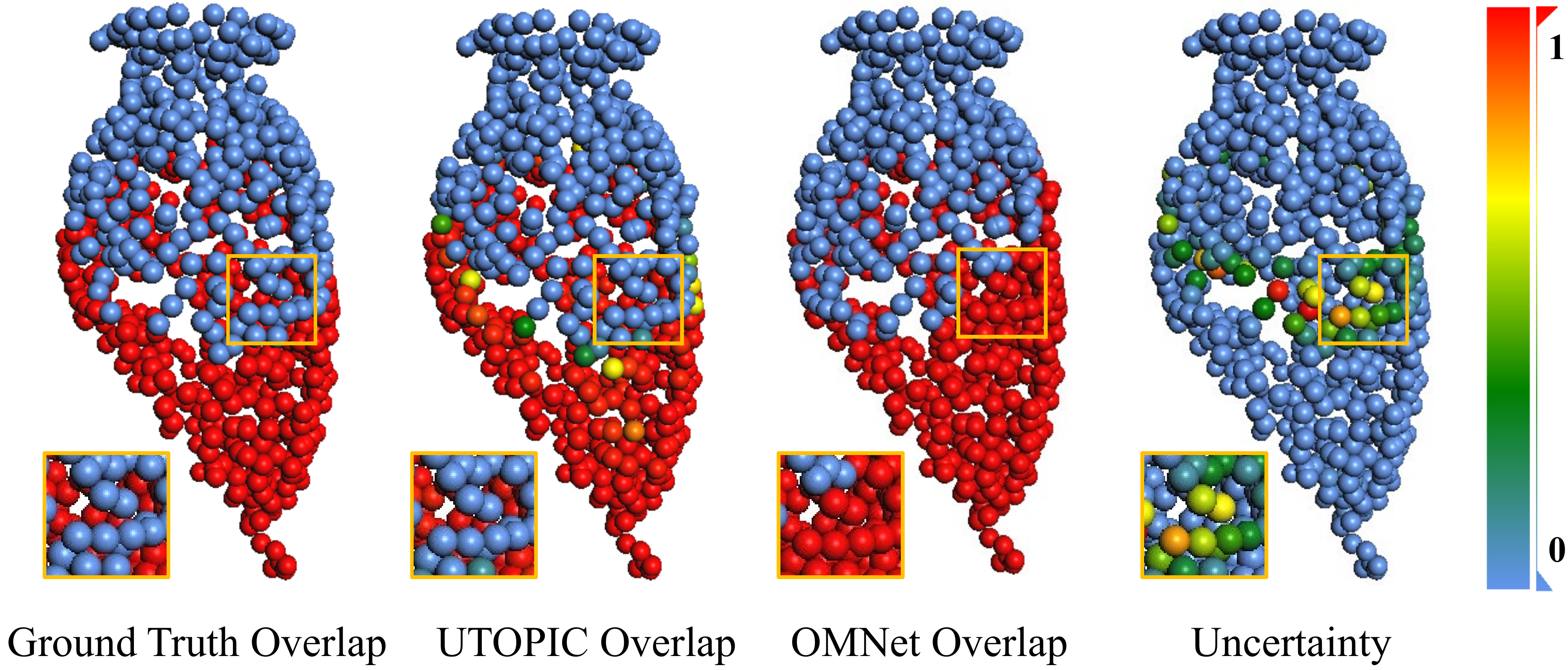}}
		\caption{Visualization of predicted overlap scores and uncertainty. We colorize both of the overlap scores and the uncertainties based on the right color scale. With the color redder, one point is more likely to be an overlapping or uncertain point.}
		\label{fig:UC}
} \end{figure}

\textbf{Visualization of predicted overlap scores and uncertainty.} We visualize the predicted overlap scores and the uncertainty in Fig.~\ref{fig:UC}. With the color redder, one point is more likely to be an overlapping or uncertain point. By comparing the predicted overlap scores between OMNet \cite{xu2021omnet} and our method and the ground-truth result, it is obvious to see that the detected overlapping regions of our method are more accurate. OMNet \cite{xu2021omnet} mistakes some ambiguous points and produces a wrong overlap mask. Besides, the redder points in the rightmost sub-figure are the points with higher uncertainty. They are almost located at the boundary between the overlapping regions and non-overlapping regions. This is consistent with our design intuition: precise uncertainty quantification facilitates better overlap prediction (Fig.~\ref{fig:UC} (b)).

\begin{table}[!htb]
\caption{Error statistics of different self-attention modules.}
\label{tab:GSA}
\centering{
\scalebox{0.655}{
\begin{tabular}{ccccccc}
\hline
UTOPIC & RMSE(\textbf{R})                       & MAE(\textbf{R})                        & RMSE(\textbf{t})                        & MAE(\textbf{t})                         & Error(\textbf{R})                      & Error(\textbf{t})                       \\ \hline
vanilla self-attention  & {0.2691} & {0.1404} & {0.00237} & {0.00136} & {0.2600} & {0.00273} \\
self-attention w/APE & {0.2667} & {0.1466} & {0.00256} & {0.00149} & {0.2718} & {0.00299} \\
self-attention w/LPE & {0.2656} & {0.1400} & {0.00226} & {0.00132} & {0.2567} & {0.00266} \\
geometry self-attention  & \textbf{0.2238} & \textbf{0.1322} & \textbf{ 0.00212} & \textbf{0.00131} & \textbf{0.2464} & \textbf{0.00264} \\
\hline
\end{tabular}}}
\end{table}

\begin{figure*}[!htb]
	\centering{{
			\includegraphics[width=1.0\linewidth]{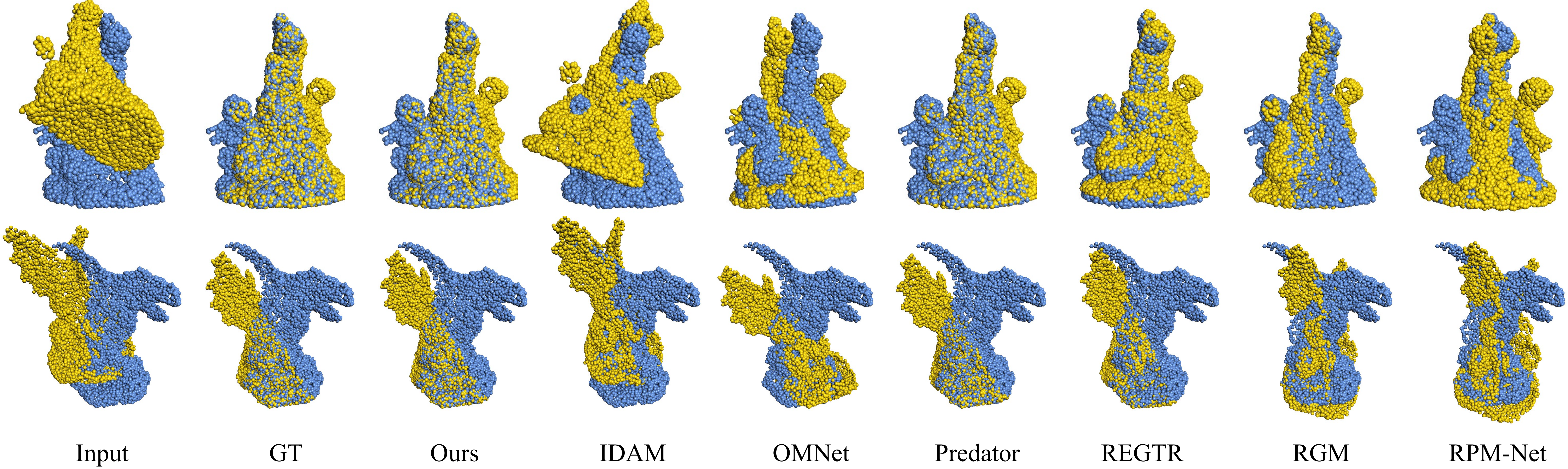}}
		\caption{Comparison of registration results on two unseen complex shapes of tiny details from PU-GAN \cite{li2019pu}.}
		\label{fig:HC}
} \end{figure*}

\textbf{Different self-attention modules.} To testify the effectiveness of the geometry self-attention, we compare four methods in Tab.~\ref{tab:GSA}: (i) vanilla self-attention \cite{yu2021cofinet}, (ii) absolute position embedding (APE) \cite{sarlin2020superglue}, (iii) learned position embedding (LPE) \cite{chen2021full}, and (iv) geometric relation embedding. We observe that the learned position embedding slightly improves the performance, while the absolute position embedding causes degradation. This is because low dimensional coordinate information without non-linear mapping is not consistent with the high dimensional features. By contrast, our approach achieves the best performance.

\begin{table}[!htb]
\caption{Analysis of different uncertainty schemes.}
\label{tab:AUS}
\centering{
\scalebox{0.66}{
\begin{tabular}{cccccccc}
\hline
UTOPIC & RMSE(\textbf{R}) & MAE(\textbf{R})                        & RMSE(\textbf{t})   & MAE(\textbf{t})  & Error(\textbf{R})  & Error(\textbf{t}) & OA\\ \hline
w/ PU, RM  & {0.3561} & {0.1562} & {0.00295} & {0.00152} & {0.2852} & {0.00309} & {82.51\%} \\
w/o U, RM & {0.3017} & {0.1547} & {0.00259} & {0.00149} & {0.2791} & {0.00300} & {91.19\%}\\
w/o RM & {0.2612} & {0.1466} & {0.00230} & {0.00142} & {0.2713} & {0.00286} & {95.31\%} \\
w/ U, RM
& \textbf{0.2238} & \textbf{0.1322} & \textbf{ 0.00212} & \textbf{0.00131} & \textbf{0.2464} & \textbf{0.00264} & \textbf{96.46\%}\\
\hline
\end{tabular}}}
\end{table}

\textbf{Analysis of different uncertainty schemes.} We conduct several experiments to elaborate the crucial role of uncertainty schemes. As shown in Tab.~\ref{tab:AUS}, we regard \emph{UTOPIC} as the baseline, and compare four variants: (i) with predicted uncertainty (\textbf{PU}) and random masking (\textbf{RM}), but the variance in Eq.~\ref{eq:uncertainty} is replaced by the predicted $\sigma_{i}$ of the Gaussian distribution, (ii) without uncertainty (\textbf{U}) and random masking (\textbf{RM}), (iii) without random masking (\textbf{RM}), (iv) with uncertainty (\textbf{U}) and random masking (\textbf{RM}). The overlap accuracy is reported as \textbf{OA} in all experiments. The first row in Tab.~\ref{tab:AUS} indicates that the registration results of this variant degrade dramatically, which shows the necessity of the computation for uncertainty in Eq.~\ref{eq:uncertainty}. This is because the predicted variance may not accurately describe the real variance of the Gaussian distribution, while the sample variance can approximate the real variance more reliably. To demonstrate the importance of uncertainty for overlap prediction as our initial motivation, we remove the uncertainty and random masking from \emph{UTOPIC} and compute the overlap accuracy. We find that the variant without uncertainty only achieves an overlap accuracy of 91.19\% on average, while the variant with uncertainty achieves 96.46\%. Comparing with \emph{UTOPIC}, the Error(\textbf{R}) of the variant without random masking degrades from 0.2464 to 0.2713 and the Error(\textbf{t}) degrades from 0.00264 to 0.00286, which demonstrates the benefits of the random masking strategy. 

\textbf{Unseen complex 3D shapes with tiny geometry details.} We additional present the visual comparisons on two unseen complex 3D shapes with tiny geometry details. The inputs are sampled with 5000 points from the dataset in \cite{li2019pu}. Notably, all compared methods are trained on the ModelNet. As shown in Fig.~\ref{fig:HC}, the sculpture and gargoyle are more geometrically complicated than the models in ModelNet40, but our approach also works well.

\begin{table}[!htb]
\caption{Timing statistics in seconds for different approaches performed on ModelNet and ModelLoNet. $N$ is the number of input points. We report the time for one pair of point clouds.}
\label{tab:TC}
\centering{
\scalebox{0.9}{
\begin{tabular}{ccc}
\hline
Method & ModelNet ($N$:717) & ModelLoNet ($N$:512)  \\\hline
IDAM \cite{li2020iterative}    & 0.07   & 0.07 \\
RPM-Net \cite{yew2020rpm} & 0.16 & 0.13     \\
OMNet \cite{xu2021omnet}   & 0.03   & 0.03  \\
RGM \cite{fu2021robust}     & 0.17  & 0.14  \\
Predator \cite{huang2021predator} & 0.29 & 0.23 \\
REGTR \cite{yew2022regtr}   & 0.05 & 0.05         \\
Ours     & 0.06 & 0.06  \\ \hline
\end{tabular}}}
\end{table}

\textbf{Timing.} Table \ref{tab:TC} reports the average running time (in seconds) of different approaches. The testing data is collected from the datasets of ModelNet and ModelLoNet. We conduct the test on a single Nvidia RTX 2080Ti with Intel Core i7-4790 @ 3.6GHz. We find that only OMNet \cite{xu2021omnet} and REGTR \cite{yew2022regtr} are faster than our method, while their performance is less satisfactory.

\section{Failure Cases and Limitations.} 
\begin{figure}[!htb]
	\centering{{
			\includegraphics[width=0.95\linewidth]{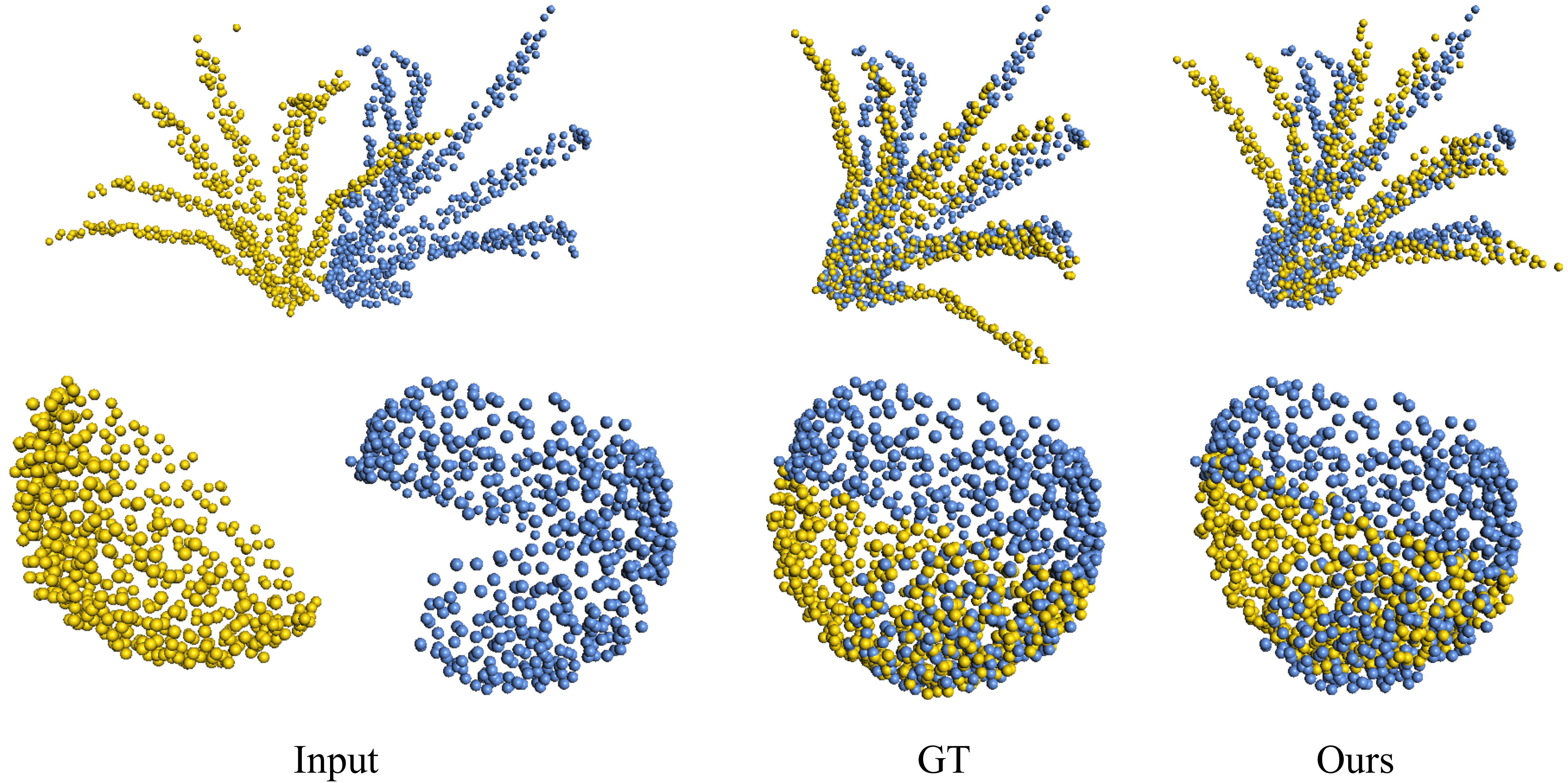}}
		\caption{Two failure examples. UTOPIC fails when: (i) the distributions of overlap regions are sparse and scattered (Row 1); (ii) the inputs lack geometrically indistinguishable features (Row 2).}
		\label{fig:FC}
} \end{figure}

Despite the promising performance of our approach, it still has some limitations. First, if the distribution of the overlapping points is sparse and scattered, it is hard for our approach to align point clouds well (see from the first row of Fig.~\ref{fig:FC}). This is because clustering points in the same overlapping areas can help generate high-certainty overlapping points.
Second, our method may fail to register the point clouds that have no distinctive geometry structures (see from the bottom row in Fig.~\ref{fig:FC}). This is because geometrically indistinguishable features lead to confusing mismatches. Finally, our approach does not conduct any downsampling operation. This hinders our method being directly adapted to large-scale point clouds.

	\section{Conclusion}
We present \emph{UTOPIC}, a novel uncertainty-aware overlap prediction network for partial point cloud registration. \emph{UTOPIC} utilizes the overlap uncertainty quantification scheme for the first time to solve the ambiguous overlap prediction problem. Through the completion decoder and geometric relation embedding, our method captures rich features of shape knowledge and transformation-invariant geometry information. Thanks to the reliable overlap scores and exact dense correspondences, \emph{UTOPIC} aligns point clouds in high accuracy, even handling low-overlap or noisy shapes. In the future, we will try to extend our method to tackle more challenging cases of large-scale (e.g., LiDAR data) point cloud registration.

    \section*{Acknowledgments}
This work was supported by the National Natural Science Foundation of China (No. 62172218, No. 62032011), the Free Exploration of Basic Research Project, Local Science and Technology Development Fund Guided by the Central Government of China (No. 2021Szvup060), the Natural Science Foundation of Guangdong Province (No. 2022A1515010170), the Innovation and Technology Fund - Midstream Research Programe for Universities from Hong Kong Innovation and Technology Commission (No. MRP/022/20X), and the General Research Fund from Hong Kong Research Grants Council (No. 15218521).
    
	\bibliographystyle{eg-alpha-doi} 
	\bibliography{Bibtex}        
	           

\end{document}